\documentclass[runningheads]{llncs}

\usepackage{amsmath}
\usepackage{amssymb}
\usepackage{graphicx}
\usepackage{pgfplots}
\pgfplotsset{compat=1.5}
\usepackage{subfigure}
\usepackage{subfig}
\usepackage{mwe}
\usepackage{algorithm}
\usepackage{algpseudocode}
\usepackage{float}
\usepackage{multirow}

\usepackage{url}
\urldef{\mailsa}\path|markku.hinkka@aalto.fi|
\urldef{\mailsb}\path|teemu.lehto@qpr.com|    
\urldef{\mailsc}\path|keijo.heljanko@helsinki.fi|
\newcommand{\keywords}[1]{\par\addvspace\baselineskip
	\noindent\keywordname\enspace\ignorespaces#1}

\usepackage{listings}
\usepackage{color}

\definecolor{dkgreen}{rgb}{0,0.6,0}
\definecolor{gray}{rgb}{0.5,0.5,0.5}
\definecolor{mauve}{rgb}{0.58,0,0.82}

\begin{document}
	
\mainmatter  

\title{Exploiting Event Log Event Attributes in RNN Based Prediction}

\titlerunning{Exploiting Event Log Event Attributes in RNN Based Prediction}

%
%
\author{Markku Hinkka\inst{1,3}\orcidID{0000-0002-3679-3677} \and Teemu Lehto\inst{1,3}\orcidID{0000-0002-1332-1853} \and Keijo Heljanko\inst{2,4}\orcidID{0000-0002-4547-2701}}
\authorrunning{Markku Hinkka, Teemu Lehto, Keijo Heljanko}

\institute{Aalto University, School of Science, Department of Computer Science, Finland
	\and
    University of Helsinki, Department of Computer Science, Finland,
	\and
	QPR Software Plc, Finland \\
	\and
	HIIT Helsinki Institute for Information Technology
	\\
	\mailsa, \mailsb, \mailsc\\
}

%
%

\toctitle{Exploiting Event Log Event Attributes in RNN Based Prediction}
\tocauthor{Markku Hinkka, Teemu Lehto, Keijo Heljanko}
\maketitle

\begin{abstract}
In predictive process analytics, current and historical process data in event logs is used to predict the future, e.g., to predict the next activity or how long a process will still require to complete. Recurrent neural networks (RNN) and its subclasses have been demonstrated to be well suited for creating prediction models. Thus far, event attributes have not been fully utilized in these models. The biggest challenge in exploiting them in prediction models is the potentially large amount of event attributes and attribute values. We present a novel clustering technique that allows for trade-offs between prediction accuracy and the time needed for model training and prediction. As an additional finding, we also find that this clustering method combined with having raw event attribute values in some cases provides even better prediction accuracy at the cost of additional time required for training and prediction. 

\end{abstract}

\section{Introduction}

\textit{Event logs} generated by systems in business processes are used in Process Mining to automatically build real-life process definitions and as-is models behind those event logs. There is a growing number of applications for predicting the properties of newly added event log cases, or process instances, based on case data imported earlier into the system~\cite{DBLP:journals/dss/EvermannRF17}\cite{DBLP:journals/corr/Francescomarino15}\cite{DBLP:conf/ssci/NavarinVPS17}\cite{DBLP:conf/caise/TaxVRD17}. The more the users start to understand their processes, the more they want to optimize them. This optimization can be facilitated by performing predictions. To be able to predict properties of new and ongoing cases, as much information as possible should be collected that is related to the event log traces and relevant to the properties to be predicted. Based on this information, a model of the system creating the event logs can be created. In our approach, the model creation is performed using supervised machine learning techniques.

In our previous work~\cite{DBLP:conf/bpm/HinkkaLHJ17} we have explored the possibility to use machine learning techniques for classification and root cause analysis for a process mining-related classification task. In the paper, experiments were performed on the efficiency of several feature selection techniques and sets of structural features (a.k.a. activity patterns) based on process paths in process mining models in the context of a classification task. One of the biggest problems with the approach is the finding of the structural features having the most impact on the classification result. E.g., whether to use only activity occurrences, transitions between two activities, activity orders, or other even more complicated types of structural features such as detecting subprocesses or repeats. For this purpose, we have proposed another approach in~\cite{DBLP:conf/bpm/HinkkaLHJ18}, where we have examined the use of recurrent neural network techniques for classification and prediction. These techniques are capable of automatically learning more complicated causal relationships between activity occurrences in \textit{activity sequences}. We have evaluated several different approaches and parameters for the recurrent neural network techniques and have compared the results with the results we collected in our work. In both the previous publications~\cite{DBLP:conf/bpm/HinkkaLHJ17}\cite{DBLP:conf/bpm/HinkkaLHJ18}, focusing on boolean -type classification tasks based on the \textit{activity sequences} only.

In this work we build on our previous work to further improve the prediction accuracy of prediction models by exploiting additional event attributes that are often available in the event logs while also taking into account the scalability of the approach to allow users to precisely specify the event attribute detail level suitable for the prediction task ahead. Our goal is to develop a technique that would allow the creation of a tool that is, based on a relatively simple set of parameters and training data, able to efficiently produce a prediction model for any case-level prediction task, such as predicting the next activity or the final duration of a running case. Fast model rebuilding is also required in order for a tool to be able to also support, e.g., interactive event and case filtering capabilities. Thus, the performance of the system under study is measured using four different metrics: Success rate, input vector length that gives a rough indication of the memory usage, the time required for training a model and the time required for performing predictions using the already trained model.

To answer these requirements, we introduce a novel method of exploiting event attributes into RNN prediction models by clustering events by their event attribute values and using the cluster labels in the RNN input vectors instead of the raw event data. This makes it easy to manage the input RNN vector size no matter how many event attributes there are in the data set. E.g., users can configure the absolute maximum length of the one-hot vector used for the event attribute data which will not be exceeded, no matter how many actual attributes the dataset has. RNN is an ideal choice for process mining-related prediction tasks since it learns temporal dynamic behavior by using its internal state to process sequences of inputs. Since predictions are usually made based on sequences of events, this makes it a more natural machine learning technique in process mining context than more traditional approaches, such as random forest and gradient boosting.

Our prediction engine source code is available in GitHub~\footnote{\url{https://github.com/mhinkka/articles}}.

The rest of this paper is structured as follows: Section~\ref{relatedwork} is a summary of the latest developments around the subject. In Section~\ref{problem}, we present the problem statement and the related concepts. Section~\ref{solution} presents our solution for the problem. In Section~\ref{testframework} we present our test framework used to test our solution. Section \ref{testsetup} describes the used datasets as well as performed prediction scenarios. Section~\ref{experiments} presents the experiments and their results validating our solution. Finally Section~\ref{conclusions} draws the final conclusions.

\section{Related Work}
\label{relatedwork}

Lately, there has been a lot of interest in the academic world on predictive process monitoring which can clearly be seen, e.g., in~\cite{DBLP:conf/bpm/Francescomarino18} where the authors have collected a survey of 55 accepted academic papers on the subject. In~\cite{DBLP:journals/corr/abs-1811-00062}, the authors have compared several approaches spanning three different research fields: Machine learning, process mining and grammar inference. As a result, they have found that overall, the techniques from machine learning field generate more accurate predictions than grammar inference and process mining fields.

In~\cite{DBLP:conf/caise/TaxVRD17} the authors used Long Short-Term Memory (LSTM) recurrent neural networks to predict the next activity and its timestamp. They use one-hot encoded activity labels and three numerical time-based features: duration between the current activity and the previous activity, time within the day and time within the week. Event attributes were not considered at all. In~\cite{DBLP:journals/dss/EvermannRF17} the authors trained LSTM networks to predict the next activity. In this case, however, network inputs are created by concatenating categorical, character string-valued event attributes and then encoding these attributes via an embedding space. They also note that this approach is feasible only because of the small number of unique values each attribute had in their test datasets. Similarly, in~\cite{DBLP:conf/enase/SchonigJAJ18}, the authors take a very similar approach based on LSTM networks, but this time also incorporate both discrete and continuous event attribute values. Discrete values are one-hot encoded, whereas continuous values are normalized using min-max normalization and added to the input vectors as single values.

In~\cite{DBLP:conf/bpm/NolleSM18} the authors use Gated Recurrent Unit (GRU) recurrent neural networks to detect \textit{anomalies} in event logs. One one-hot encoded vector is created for activity labels and one for each of the included string-valued event attributes. These vectors are then concatenated in a similar fashion to our solution into one vector representing one event, which is then given as input to the network. We use this approach for benchmarking our own clustering-based approach (labeled as \textit{Raw} feature in the text below). The system proposed in their paper is able to predict both the next activity and the next values of event attributes. Specifically, it does not take case attributes and temporal attributes into account.

In~\cite{quteprints96732} the authors train a RNN to predict the most likely future activity sequence of a running process based only on the sequence of activity labels. Similarly our earlier publication~\cite{DBLP:conf/bpm/HinkkaLHJ17} used sequences of activity labels to train a LSTM network to perform a boolean classification of cases. 

Also, process mining models obtained using process mining techniques themselves can be used as a model for prediction. In~\cite{DBLP:conf/adbis/BernardA19} the authors first generate a process tree using the inductive miner algorithm, after which this process tree is used to predict the future path of ongoing cases. This approach does not take any additional event- or case attributes into account.

None of the mentioned earlier works present a solution that is scalable for datasets having lots of event- or case attributes and unique attribute values.

\section{Problem}
\label{problem}

Using RNN to perform case-level predictions on event logs has lately been studied a lot. However, there has not been any scalable approach to handling event attributes in the RNN setting. Instead, e.g., in~\cite{DBLP:conf/bpm/NolleSM18} authors used separate one-hot encoded vector for each attribute value. Having this kind of an approach when you have, e.g., 10 different attributes, each having 10 unique values would already require a vector of 100 elements to be added as input for every event. The longer the input vectors become, the more time and memory it gets for the model to create accurate models from them. This increases the time and memory required to use the model for predictions. 

\section{Solution}
\label{solution}

Since in addition to having activity labels in the input vectors, we need to store also event attribute-related information, we decided to include several feature types into the input vectors of the RNN. Input vectors are formatted as shown in Table~\ref{table:input-vector-format}, where each column represents one feature vector element $f_{ab}$, where $a$ is the index of the feature and $b$ is the index of the element of that feature. In the table, $n$ represents the number of feature types used in the feature vector and $m_k$ represents the number of elements required in the input vector for feature type $k$. Thus, each feature type produces one or more numeric elements into the input vector, which are then concatenated together into one actual input vector passed to RNN both in training and in prediction phases. Table~\ref{table:example-input-vector} shows an example input vector having three different feature types: activity label, raw event attribute values (only single event attribute named $\mathit{food}$ having four unique values) and the event attribute cluster where clustering has been performed separately for each unique activity. 

For this paper, we encoded only event activity labels and event attributes into the input vectors. However, this mechanism can easily incorporate also other types of features not described here. The only requirement for added features is that it needs to be able to be encoded into a numeric vector as shown in Table~\ref{table:input-vector-format} whose length must be the same for each event.

\begin{table*}[!t]
	\centering
	\caption{Feature input vector structure}
	\label{table:input-vector-format}
	\begin{tabular}{|c|c|c|c|c|c|c|c|c|c|c|}
		\hline
		$f_{11}$ & $f_{12}$ & ... & $f_{1m_1}$ & $f_{21}$ & ... & $f_{2m_2}$ & ... & $f_{n1}$ & ... & $f_{nm_n}$  \\ \hline
	\end{tabular}
\end{table*}

\begin{table*}[!t]
	\centering
	\caption{Feature input vector example content}
	\label{table:example-input-vector}
	\begin{tabular}{|c|c|c|c|c|c|c|c|c|c|c|}
		\hline
		$row$ & $activity_{eat}$ & $activity_{drink}$ & $food_{salad}$ & $food_{pizza}$ & $food_{water}$ & $food_{soda}$ & $cluster_1$ & $cluster_2$ \\ \hline
		1 & 1 & 0 & 1 & 0 & 0 & 0 & 1 & 0 \\
		2 & 0 & 1 & 0 & 0 & 1 & 0 & 1 & 0 \\
		3 & 1 & 0 & 0 & 1 & 0 & 0 & 0 & 1 \\
		4 & 1 & 0 & 0 & 1 & 0 & 0 & 0 & 1 \\
		5 & 0 & 1 & 0 & 0 & 0 & 1 & 0 & 1 \\
		\hline
	\end{tabular}
\end{table*}

\subsection{Event Attributes}
\label{eventattributes}

Our primary solution for incorporating information in event attributes into input vectors is to cluster all the event attribute values in the training set and then use a one-hot encoded cluster identifier to represent all the attribute values of the element. The used clustering algorithm must be such that it tries to automatically find the optimal number of clusters for the given data set within the range of 0 to N clusters, where N can be configured by the user. By changing N, the user can easily configure the maximum length of the one-hot -vector as well as the precision of how detailed attribute information will be tracked. For this paper, we experimented with a slightly modified version of Xmeans -algorithm~\cite{DBLP:conf/icml/PellegM00}. Another option could have been a method where silhouette scoring~\cite{rousseeuw1987silhouettes} is used to determine the optimal number of clusters for k-means~\cite{hartigan1979algorithm}, but based on our tests, this approach did not perform fast enough to be applied as our selected approach.

It is very common that different activities get processed by different resources yielding a completely different set of possible attribute values. E.g., different departments in a hospital have different people, materials and processes. Also, in the example feature vector shown in Table~\ref{table:example-input-vector}, $\mathit{food}$ -event attribute has completely different set of possible values depending on the $activity$ since it is forbidden by, e.g., the external system to not allow activity of type \textit{eat} to have \textit{food} event attribute value of \textit{water}. If we cluster all the event attributes using single clustering, we would easily lose this activity type-specific information.

In order to retain this activity-specific information, we used separate clustering for each unique activity type. All the event attribute clusters are encoded into one one-hot encoded vector representing only the resulting cluster label for that event, no matter what its activity is. This is shown in the example table as $cluster_N$, which represents the row having $N$ as a clustering label. E.g., in the example case, $cluster_1$ is 1 in both rows 1 and 2. However, row 1 is in that cluster because it is in the 1st cluster of the $activity_{eat}$ activity, whereas row 2 is in that cluster because it is in the 1st cluster of the $activity_{drink}$ activity. Thus, in order to identify the actual cluster, one would require both the activity label and the cluster label. For RNN to be able to properly learn about the actual event attribute values, it needs to be given both the activity label and the cluster label in the input vector. Below, this approach is labeled as \textit{ClustN}, where \textit{N} is the maximum cluster count.

For benchmarking, we also experimented with a \textit{raw} implementation where event attributes were used so that every event attribute is encoded into its own one-hot encoded vector and then concatenated into the actual input vectors. This method is lossless since every unique event attribute value has its own indicator in the input vector. Below, this approach is referred to as \textit{Raw}. Finally, we experimented also using both \textit{Raw} and \textit{Clustered} event attribute values. Below, this approach is referred to as \textit{BothN}, where \textit{N} is the maximum cluster count.

\subsection{Formal problem definition}
\label{formaldefinition}

Basically, the problem we are solving in this paper is that we encode as much information as possible from event log event attributes into a single numeric RNN input vector whose length is user-configurable. In this section, we will present the formal definitions required to describe our clustering-based approach for solving this problem. We will build our formal definitions on the basis of the definitions given in Process Mining ~\cite{van2011process} -book Chapter 5 describing the basic concepts in process mining. 

First, we define relationships between event log and events as follows:
\begin{definition}
	\label{def:eventlogevent} Event log $L$ is a set of cases $c_i \in L$. Each case $c_i$ is a sequence of events $c_i = \langle e_1, e_2, ..., e_n \rangle$. We denote $e \in L$ iff $\exists c_i = \langle e_1, e_2, ..., e_n \rangle \in L$, such that $e_j = e$, for some $1 \leq j \leq n$.
\end{definition}

Next, we define a function for accessing event attributes of an event:

\begin{definition}
	\label{def:eventattributes} Let $\#_n : e \mapsto v$, where $(n \in AN) \land (e \in L)$, and $AN$ is the finite set of all the attribute labels for the events in the given event log. $v$ is the value of that attribute and can be of any arbitrary type.
\end{definition}

Using this function, we can refer to any of the \textit{standard attributes} (in this paper, only \textit{activity}, \textit{time} and \textit{caseid} are considered as standard attributes), as well as event log-specific attributes. For this paper, we need to define these attribute label sets as follows:

\begin{definition}
	\label{def:standardattributes} Let $AN_{std} \subseteq AN$, be the set of standard attribute labels: \\ $\langle activity, time, caseid \rangle$. Also, let $(AN_L \cap AN_{std} = \emptyset) \land (AN_{std} \cup AN_L = AN)$, where $AN_L$ is the set of event attributes in event log $L$, which are not part of standard attributes.
\end{definition}

The standard attributes listed above have the following meanings:

\begin{definition}
	\label{def:standardeventattributes}
	$\#_{activity}(e)$ is the activity label associated with the event $e$. This describes what has occurred. Activity labels in this paper are considered to be textual descriptions of the performed task.\\
	$\#_{time}(e)$ is the timestamp of the event $e$. This describes when something has occurred. In this paper, timestamps include both time and date of the occurred event.\\
	$\#_{caseid}(e)$ is the identifier of the case associated with the event $e$. This describes the (case) context of the occurrence. Case identifiers in this paper are considered to be short textual or numeric identifiers identifying the case.\\
\end{definition}

Using these, we can formally define a set of all the activity labels in the event log as follows:

\begin{definition}
	Let $\mathcal{A}_L$ be the set of activity labels so that $\forall e \in L, \#_{activity}(e) \in \mathcal{A}_L$.
\end{definition}

Next, we split the event log into two disjoint sets: Training set and test set. Formally this can be expressed as:

\begin{definition}
	\label{def:trainingset} Training set is $L_{tr} \subset L$, where $L_{tr} \neq \emptyset$. Similarly, test set is $L_{t} \subset L$, where $(L_{t} \neq \emptyset) \land (L_{t} \cap L_{tr} = \emptyset) \land (L_{t} \cup L_{tr} = L)$. 
\end{definition}

We also formally define separate subsets of activity labels for both the test and the training set as follows:

\begin{definition}
	$\mathcal{A}_{tr} \subseteq \mathcal{A}_L$ is used to denote all the activity labels available in the training set, whereas $\mathcal{A}_{t} \subseteq \mathcal{A}_L$ denotes those in the test set. 
\end{definition}

Similarly, we denote sets of available attributes in both the training set and the test set as follows:

\begin{definition}
	$AN_{tr} \subseteq AN_L$ is used to denote all the attribute names available in the training set, whereas $AN_{t} \subseteq AN_L$ denotes those in the test set.
\end{definition}

Next, we define a function used to concatenate multiple vectors to each other.

\begin{definition}
	\label{def:concat} Let $\mathit{concat} : \langle X_0, ..., X_n \rangle \mapsto Y$ be a function that returns a single numeric vector $Y$ consisting of the concatenated contents of numeric vectors $X_0, ..., X_n$ in the specified order.
\end{definition}

Then we define a function that maps each unique value into unique integer value as follows:

\begin{definition}
	\label{def:codify} Let $\mathit{codify} : x \times X \mapsto y$, where $X$ is a finite set of all the possible values for $x \in X$, and $y \in {1, ..., |X|}$ is the value $x$ being mapped bijectively into an integer. Thus, $\mathit{codify}$ creates a bijection between values and an integer representation of that value.
\end{definition}

Now we can give a formal definition for the one-hot encoding function as follows:

\begin{definition}
	\label{def:onehot} Let $\mathit{onehot} : x \times X \mapsto Y$, where $X$ is a finite set of all the possible values and $x \in X$ is the value being encoded, be an one\-hot encoding function that transforms $x$ into a numeric vector $Y$ of length $|X|$. Vector $Y = \langle y_1, ..., y_{|X|} \rangle$, where \[
	y_k = 
	\begin{cases}
	1,& \text{iff } k = \mathit{codify}(x, X)\\
	0,& \text{otherwise}
	\end{cases}
	\]
	Thus, every unique value in $X$ returns an unique numeric vector. 
\end{definition}

An example of a one-hot encoding process is shown in Table~\ref{table:one-hot-example}, where the set of all possible values $X$ is $\langle a, b, c, d \rangle$, and we are applying $\mathit{onehot}$ function to each of the four possible values separately.

\begin{table*}[!t]
	\centering
	\caption{One-hot encoding example}
	\label{table:one-hot-example}
	\begin{tabular}{|c|c|c|}
		\hline
		Original & After $\mathit{codify}$ & After $\mathit{onehot}$  \\ \hline
		$\langle a, b, c, d \rangle$ & $\langle 1, 2, 3, 4 \rangle$ & $\langle \langle 1, 0, 0, 0 \rangle, \langle 0, 1, 0, 0 \rangle, \langle 0, 0, 1, 0 \rangle, \langle 0, 0, 0, 1 \rangle \rangle$ \\ \hline
	\end{tabular}
\end{table*}

Next, since we need to be able to iterate through all the attributes, we need to specify an unambiguous way to map iteration index to an attribute name. For this purpose we define the following function:

\begin{definition}
	\label{def:attributeiteration} Let $\mathit{attname} : n \mapsto an$, where $an \in AN_L$, and $n \in {1, ..., |AN_L|}$ be a bijective function for mapping an positive integer iteration index to an attribute name.
\end{definition}

Using the previous definition, we can refer to $n$th event attribute of $e$ by writing $\#_{\mathit{attname}(n)}(e)$. Next, we define a method for creating subsets of events in a way that each subset will have all the events having one specific activity label. Formally we express these sets as:

\begin{definition}
	\label{def:activitybuckets} Let $B_{act}$, where $act \in \mathcal{A}_{tr}$ be the set of events $e \in L$ such that $\#_{activity}(e) = act$. 
\end{definition}

Next, we will define a function for retrieving a set of all attribute values of a given set of events:
\begin{definition}
	\label{def:eventsetattributes} Let $\#_n : \langle e_1, ..., e_n \rangle \mapsto Y$
	, where $e_i \in L \forall (1 \leq i \leq n)$. This function returns a set of all the attribute values of attribute having index $n$ for given events.
\end{definition}

Using these definitions, the set of all the attribute values of all the events having a specific activity label $act$ can be referred to using $\#_{\mathit{attname}(n)}(B_{act})$, which yields a set of attribute values of attribute having index $n$ for all the events having activity label $act$. Next, In order to perform clustering for events in activity buckets, we need to first one-hot encode event attribute values.

\begin{definition}
	\label{def:clusteringvectorization} Let $\mathit{onehot}_{B_{act}} : e \times n \mapsto Y$, where $e \in L$, $\mathit{act} \in \mathcal{A}_{tr}$, and $n \in {1, ..., |AN_L|}$, be a function that performs one-hot encoding for attribute having attribute iteration index of $n$ for event $e$. Thus, \\ $Y = \mathit{onehot}(\#_{\mathit{attname}(n)}(e), \#_{\mathit{attname}(n)}(B_{\#_{activity}(e)}))$.
\end{definition}
	
Using this definition, we can transform all event attribute values into a single numeric vector using the following additional function definition.

\begin{definition}
	\label{def:onehot-alleventattributes} One-hot encoding function for all event attributes of given event:
\begin{align*}
	onehot_{\#} : e \mapsto \mathit{concat}(&\mathit{onehot}_{B_{\#_{activity}(e)}}(e, 1), \\
	&..., \\
	&\mathit{onehot}_{B_{\#_{activity}(e)}}(e, |AN_L|))
\end{align*}
, where $e \in L$.
\end{definition}

Next, we define the actual clustering function that uses the number vectors created by $\mathit{onehot}_{\#}$ function as follows.

\begin{definition}
	\label{def:clusteringfunction} $\#_{cluster} : e \times max_{cc} \mapsto y$, where $e \in L_{tr}$, and $max_{cc}$ denotes the configured maximum cluster count, and $1 \leq y \leq max_{cc}$, and $y \in \mathbb{Z}$, denotes the assigned cluster label among the set of all the possible cluster labels. Clustering is performed separately for each activity label $B_{act}$, where $act \in \mathcal{A}_{tr}$ using suitable clustering algorithm. Every clustered event is translated into clustering input vector using $\mathit{onehot}_{\#}(e)$ function. These input vectors are then used as data points for the clustering algorithm.
\end{definition}

Thus, every activity will have its own independent clustering having $max_{cc}$ as the maximum cluster count. In the testing phase, the clusterings created from the training data will be used to fit the input vectors created from the events in the testing data. 

Finally, we can specify the vector used as input vector in RNN training as follows:

\begin{definition}
	\label{def:traininginputvector} Generating input vector for one event in training is performed using:
\begin{align*}
	inputvector : e \times max_{cc} \mapsto \mathit{concat}(&\mathit{onehot}(\#_{activity}(e), ACT_{tr}), \\
	&\mathit{onehot}(\#_{cluster}(e, max_{cc}), cl))
\end{align*}
, where $e \in L_{tr}$, and $cl = \langle 1, ..., n\rangle$, where $n \leq max_{cc}$ represents the set of all the possible cluster labels. The value of $n \in \mathbb{Z}$ depends on the clustering algorithm and represents the actual maximum number of clusters that were created for event attribute data of any single activity in $L_{tr}$.
\end{definition}

These input vectors are then passed to the RNN training as ordered sequences of event input vectors, where each sequence represents all the events of a single case in the $L_{tr}$ in the order determined by their ascending timestamps.

As a result, when training, every event is preprocessed by performing the input vector generation using $inputvector$ function. At the testing phase, the same encoding functions are used, however, if the event used in testing has some attributes, attribute values or activities that were not part of the training data set, those will just be ignored. Also, event attribute clustering in the testing phase is performed using the clustering models created in the training phase. Thus, in order to store the trained model, also all the trained clustering models must be stored.

\section{Test Framework}
\label{testframework}

We have performed our test runs using an extended Python-based prediction engine that was used in our earlier work~\cite{DBLP:conf/bpm/HinkkaLHJ17}. The engine is still capable of supporting most of the hyperparameters that we experimented with in our earlier work, such as used RNN unit type, number of RNN layers and the used batch size. The prediction engine we built for this work takes a single JSON configuration file as input and outputs test result rows into a CSV file. 

Tests were performed using a commonly used 3-fold cross-validation technique to measure the generalization characteristics of the trained models. In 3-fold cross-validation, the input data is split into three subsets of equal size. Each of the subsets is tested one by one against models trained using the other two subsets.

\subsection{Training}
\label{predictionengine-training}

Training begins by loading the event log data contained in the two of the three event log subsections.  After this, the event log is split into actual training data and validation data that used to find the best performing model out of all the model states during all the test iterations. For this, we picked 75\% of the cases for the training and the rest for the validation dataset. After this, we initialize event attribute clusters as described in Section~\ref{eventattributes}. 

The actual prediction model and the data used to generate the actual input vectors is performed next. This data initialization involves splitting cases into prefixes and also taking a random sample of the actual available data if the amount of data exceeds the configured maximum amount of prefixes. To avoid running out of memory during any of our tests, these limits were set to 75000 for training data and 25000 for validation data. We also had to filter out all the cases having more than 100 events.

Finally, after the model is initialized, we start the actual training in which we concatenate all the requested feature vectors as well as the expected outcome into the RNN model repeatedly for the whole training set until 100 test iterations have passed. The number of actual epochs trained in each iteration is configurable. In our experiments, the total number of epochs was set to be 10. After every test iteration, the model is validated against the validation set. To improve validation performance, if the size of the validation set is larger than a separately specified limit (10000), a random sample of the whole validation set is used. These test results, including additional status and timing-related information, are written into resulting test result CSV file. If the prediction accuracy of the model against the validation set is found to be better than the accuracy of any of the models found thus far, then the network state is stored for that model. Finally, after all the training, the model having the best validation test accuracy is picked as the actual result of the model training.

\subsection{Testing}
\label{predictionengine-testing}

In the testing phase, the third subset of cross-validation folding is tested against the model built in the previous step. After initializing the event log following similar steps as in the training phase, the model is asked for a prediction for each input vector built from the test data. To prevent running out of memory and to ensure tests are not taking an exceedingly long time to run, we limited the number of final test traces to 100000 traces and used random sampling when needed. The prediction result accuracy, as well as other required statistics, are written to the resulting CSV file.

\section{Test Setup}
\label{testsetup}

We performed our tests using several different data sets. Some details of the used data sets can be found in the Table~\ref{table:testdatasets}. The table lists the number of cases, activities, events and event attributes for each event log. \textit{\# Unique values} column shows the sum of all the unique attribute values for each of the selected attributes.

\begin{table}[!t]
	\centering
	\caption{Used Event logs and their relevant statistics}
	\label{table:testdatasets}
	\begin{tabular}{|l|c|c|c|c|c|}
		\hline
		Event log & \# Cases & \# Activities & \# Events & \# Attributes & \# Unique values \\ \hline
		BPIC12\footnotemark & 13087 & 24 & 262200 & 1 & 3 \\
		BPIC13, incidents\footnotemark & 7554 & 13 & 65533 & 8 & 2890 \\
		BPIC14\footnotemark & 46616 & 39 & 466737 & 1 & 242 \\
		BPIC17\footnotemark & 31509 & 26 & 1202267 & 4 & 164 \\
		BPIC18\footnotemark & 43809 & 41 & 2514266 & 5 & 360 \\
		\hline
	\end{tabular}
\end{table}

\addtocounter{footnote}{-5}
\stepcounter{footnote}\footnotetext{\url{https://doi.org/10.4121/uuid:3926db30-f712-4394-aebc-75976070e91f}}
\stepcounter{footnote}\footnotetext{\url{https://doi.org/10.4121/uuid:500573e6-accc-4b0c-9576-aa5468b10cee}}
\stepcounter{footnote}\footnotetext{\url{https://doi.org/10.4121/uuid:c3e5d162-0cfd-4bb0-bd82-af5268819c35}}
\stepcounter{footnote}\footnotetext{\url{https://doi.org/10.4121/uuid:5f3067df-f10b-45da-b98b-86ae4c7a310b}}
\stepcounter{footnote}\footnotetext{\url{https://doi.org/10.4121/uuid:3301445f-95e8-4ff0-98a4-901f1f204972}}

The criteria for selecting or not selecting an event attribute in a model were based on the maximum usage of any unique value that the attribute has in the event log. If a value of an attribute was used in more than 4\% of all the events in the event log, then that attribute was included in the test. Besides, we did not select any attributes that had just one unique attribute value. Names of all the selected event attributes are listed in the Table~\ref{table:attributenames}.

\begin{table}[!t]
	\centering
	\caption{Included event attributes}
	\label{table:attributenames}
	\begin{tabular}{|l|l|}
		\hline
		Event log & Attribute names \\ \hline
		BPIC12 & lifecycle:transition \\
		\hline
		\multirow{8}{*}{BPIC13, incidents} & impact \\
		& lifecycle:transition \\
		& org:group \\
		& org:resource \\
		& organization country \\
		& organization involved \\
		& product \\
		& resource country \\
		\hline
		BPIC14 & Assignment Group \\
		\hline
		\multirow{4}{*}{BPIC17} & Action \\
		& EventOrigin \\
		& lifecycle:transition \\
		& org:resource \\
		\hline
		\multirow{5}{*}{BPIC18} & activity \\
		& doctype \\
		& note \\
		& org:resource \\
		& subprocess \\
		\hline
	\end{tabular}
\end{table}

For each dataset, we performed the next activity prediction where we wanted to predict the next activity of any ongoing case. This was accomplished by splitting every input case into possibly multiple \textit{virtual} cases depending on the number of events the case had. If the length of the case was shorter than 4, the whole case was ignored. If the length was equal or higher, then a separate \textit{virtual} case was created for all prefixes at least of length 4. Thus, for a case of length 6, 3 cases were created: One with length 4, one with 5 and one with 6. For all these prefixes, the next activity label was used as the expected outcome. For the full-length case, the expected outcome was a special \textit{finished}-token. 

\section{Experiments}
\label{experiments}

For experiments, we have used the same system that we used already in our previous work~\cite{DBLP:conf/bpm/HinkkaLHJ17}. The system had Windows 10 operating system and its hardware consisted of 3.5 GHz Intel Core i5-6600K CPU with 32 GB of main memory and NVIDIA GeForce GTX 960 GPU having 4 GB of memory. Out of those 4 GB, we reserved 3 GB for the tests. The testing framework was built on the test system using the Python programming language. The actual recurrent neural networks were built using Lasagne~\footnote{https://lasagne.readthedocs.io/} library that works on top of Theano~\footnote{http://deeplearning.net/software/theano/}. Theano was configured to use GPU via CUDA for expression evaluation. 

We used one layer GRU~\cite{DBLP:conf/ssst/ChoMBB14} as the RNN type. Adam~\cite{DBLP:journals/corr/KingmaB14} was used as gradient descent optimizer with parameters of $beta_1=0.9$ and $beta_2=0.999$. 256 was used as the hidden dimension size as well as the mini-batch size and 0.01 as the learning rate. Even though it is quite probable that more accurate results could have been achieved by selecting, e.g., different hidden dimension sizes and learning rates depending on the size of the input vectors, we decided to use the fixed values. This decision was made in order to make the interpretation of the test results easier by minimizing the number of variables affecting the test results.

We performed next activity predictions using all the four combinations of features, five data sets and three different maximum cluster counts: 20, 40, and 80 clusters. The results of these runs are shown in Table~\ref{table:next-activity-by-dataset}. In the table, \textit{Features} -column shows the used set of features. \textit{S.rate} shows the achieved prediction success rate. \textit{In.v.s.} shows the size of the input vector. This column can be used to give some kind of indication on the memory usage of using that configuration. Finally, \textit{Tra.t.} and \textit{Pred.t.} columns tell us the time required for performing the training and the prediction for all the cases in the test dataset. In both cases, this time includes the time for setting up the neural network, clusterings and preparing the dataset from JSON format. Sample standard deviation has been included in both \textit{S.rate} and \textit{Tra.t} in parentheses to indicate how spread out the measurements are within all the three test runs. Each row in the table represents three cross-validation runs with a unique combination of dataset and feature that was tested. Rows having the best prediction accuracy within a dataset are shown using bold font. \textit{None} -feature represents the case in which there were no event attribute information at all in the input vector, \textit{ClustN} represents a test with one-hot encoded cluster labels of event attributes clustered into maximum of \textit{N} clusters, \textit{Raw} represents having all one-hot encoded attribute values individually in the input vector, and finally, \textit{BothN} represents having both one-hot encoded attribute values and one-hot encoded cluster labels in the input vector.

\begin{table}[!t]
	\centering
	\caption{Statistics of next activity prediction using different sets of input features}
	\label{table:next-activity-by-dataset}
	\begin{tabular}{|l|c|c|c|c|c|}
		\hline
		Dataset & Features & S.rate ($\sigma$) & In.v.s. & Tra.t. ($\sigma$) & Pred.t. \\ \hline
		\multirow{8}{*}{BPIC12} & None & 85.8\% (0.3\%) & 25.7 & 489.0s (7.0s) & 35.1s  \\
		& Clust20 & 86.0\% (0.4\%) & 30.0 & 500.6s (2.5s) & 31.6s \\
		& Clust40 & 85.8\% (0.3\%) & 30.0 & 499.7s (1.3s) & 31.9s \\
		& Clust80 & 86.2\% (0.1\%) & 30.0 & 502.1s (2.3s) & 7.5s \\
		& Raw & 85.9\% (0.3\%) & 29 & 504.3s (0.5s) & 38.9s \\
		& Both20 & 86.0\% (0.2\%) & 33 & 515.3s (2.6s) & 40.4s \\
		& Both40 & 86.0\% (0.4\%) & 33 & 517.7s (3.6s) & 40.4s \\
		& \textbf{Both80} & \textbf{86.3\% (0.1\%)} & \textbf{33} & \textbf{518.2s (4.0s)} & \textbf{40.7s} \\
		\hline
		\multirow{8}{*}{BPIC13} & None & 62.9\% (0.3\%) & 13.7 & 165.6s (21.2s) & 3.5s  \\
		& Clust20 & 66.8\% (0.3\%) & 34.7 & 188.0s (22.4s) & 4.7s \\
		& Clust40 & 67.2\% (0.7\%) & 54.7 & 214.8s (3.1s) & 5.4s \\
		& Clust80 & 67.0\% (0.6\%) & 94.7 & 258.4s (4.7s) & 6.0s \\
		& Raw & 68.2\% (1.1\%) & 2353.7 & 2611.7s (44.7s) & 74.8s \\
		& \textbf{Both20} & \textbf{69.1\% (0.6\%)} & \textbf{2359.3} & \textbf{2464.6s (309.0s)} & \textbf{94.4s} \\
		& Both40 & 68.9\% (0.5\%) & 2395.7 & 2687.1s (227.3s) & 106.6s \\
		& Both80 & 68.4\% (0.7\%) & 2429.3 & 2821.8s (33.5s) & 194.3s \\
		\hline
		\multirow{8}{*}{BPIC14} & None & 37.8\% (1.5\%) & 40.3 & 488.1s (5.3s) & 36.1s  \\
		& Clust20 & 39.9\% (0.5\%) & 61.7 & 523.3s (3.5s) & 40.4s \\
		& Clust40 & 40.0\% (0.3\%) & 80.3 & 553.5s (3.8s) & 43.6s \\
		& Clust80 & 40.2\% (0.1\%) & 84.7 & 556.8s (10.5s) & 43.6s \\
		& Raw & 39.7\% (1.4\%) & 272.0 & 825.7s (2.8s) & 68.0s \\
		& Both20 & 40.6\% (0.6\%) & 292.3 & 907.1s (7.5s) & 78.6s \\
		& \textbf{Both40} & \textbf{40.6\% (0.6\%)} & \textbf{309.3} & \textbf{943.3s (10.6s)} & \textbf{82.0s} \\
		& Both80 & 37.3\% (4.2\%) & 305.0 & 935.1s (26.9s) & 156.7s \\
		\hline
		\multirow{8}{*}{BPIC17} & None & 86.4\% (0.4\%) & 27.7 & 518.7s (2.8s) & 107.7s  \\
		& \textbf{Clust20} & \textbf{90.8\% (0.3\%)} & \textbf{48.7} & \textbf{556.3s (3.7s)} & \textbf{132.4s} \\
		& Clust40 & 90.2\% (1.4\%) & 68.3 & 637.5s (58.3s) & 143.9s \\
		& Clust80 & 90.2\% (0.4\%) & 108.7 & 647.3s (3.7s) & 142.8s \\
		& Raw & 89.9\% (0.5\%) & 190 & 816.4s (5.2s) & 164.9s \\
		& Both20 & 89.9\% (0.5\%) & 211.0 & 867.8s (3.5s) & 188.0s \\
		& Both40 & 90.2\% (0.3\%) & 230.3 & 910.9s (19.3s) & 193.7s \\
		& Both80 & 89.6\% (0.6\%) & 271.3 & 986.5 (4.4s) & 197.7s \\
		\hline
		\multirow{8}{*}{BPIC18} & None & 71.3\% (9.3\%) & 43 & 516.0s (9.5s) & 197.0s  \\
		& Clust20 & 79.0\% (0.9\%) & 64.0 & 588.7s (13.7s) & 268.7s \\
		& \textbf{Clust40} & \textbf{79.9\% (0.2\%)} & \textbf{84.0} & \textbf{628.4s (2.8s)} & \textbf{286.1s} \\
		& Clust80 & 79.5\% (0.1\%) & 124.0 & 701.3s (7.4s) & 306.9s \\
		& Raw & 79.3\% (0.4\%) & 349.7 & 1173.7s (83.1s) & 381.2s \\
		& Both20 & 79.7\% (0.5\%) & 377.7 & 1213.1s (48.1s) & 463.2s \\
		& Both40 & 79.9\% (0.5\%) & 401.0 & 1301.9s (82.9s) & 540.3s \\
		& Both80 & 79.3\% (0.5\%) & 425.7 & 1405.4s (87.2s) & 619.9s \\
		\hline
	\end{tabular}
\end{table}

\begin{figure*}[!htbp]
	\begin{minipage}{\linewidth}
		\centering
		\begin{minipage}{0.45\linewidth}
			\begin{figure}[H]
				\includegraphics[trim={2cm 10.2cm 2cm 11.6cm},clip,width=0.9\linewidth]{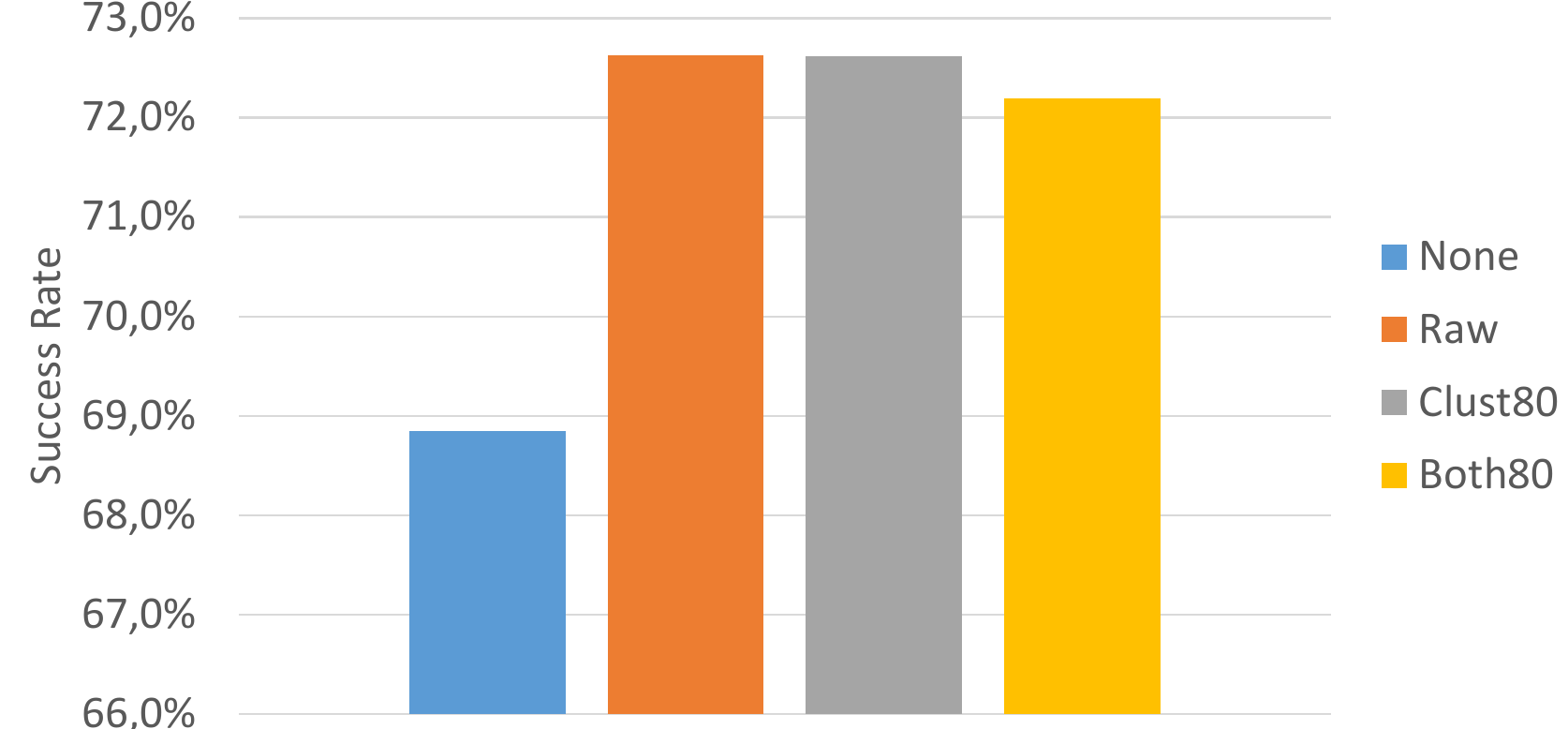}
				\caption[]{Average prediction success rate over all the datasets}
				\label{figure:next-activity-success-rate}
			\end{figure}
		\end{minipage}
		\hspace{0.05\linewidth}
		\begin{minipage}{0.45\linewidth}
			\begin{figure}[H]
				\includegraphics[trim={2cm 10.2cm 2cm 11.6cm},clip,width=0.9\linewidth]{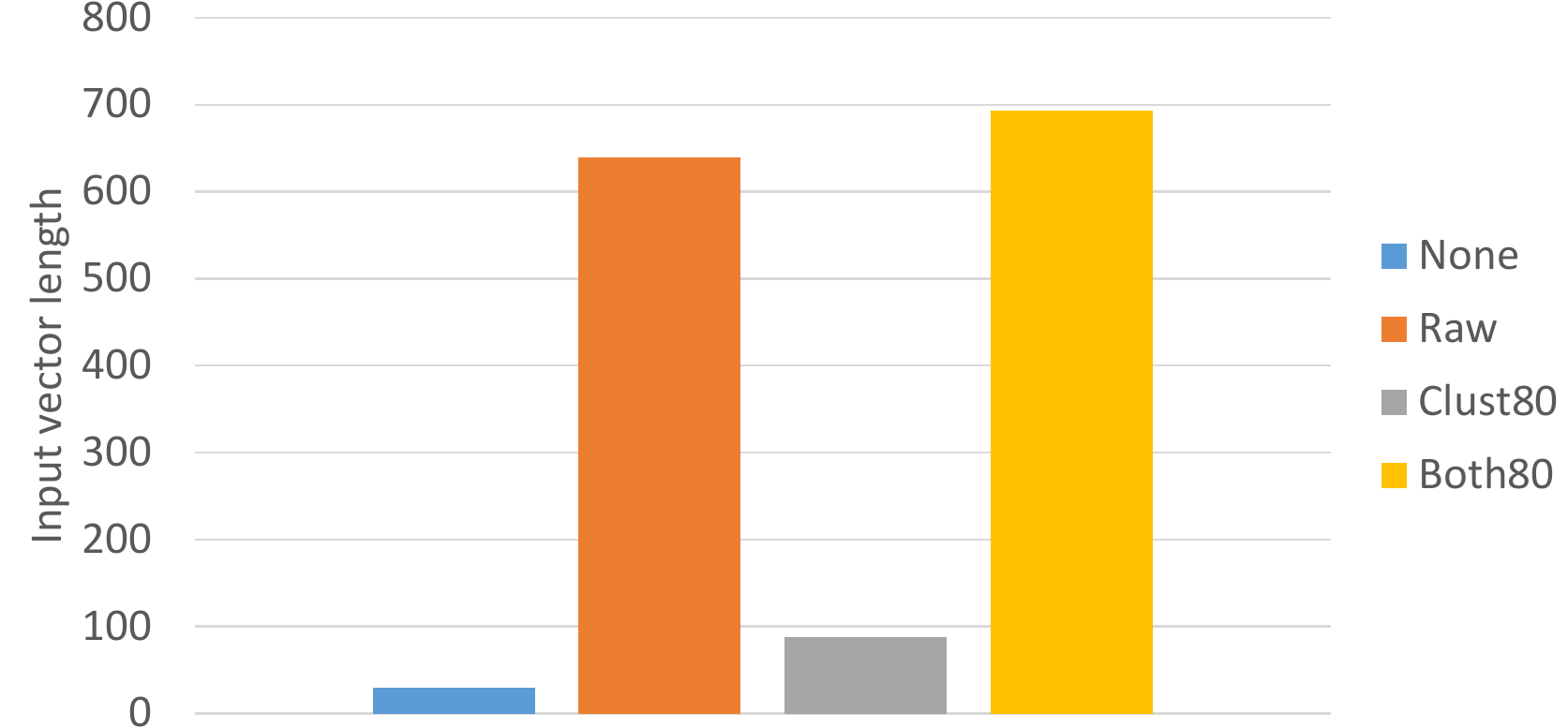}
				\caption[]{Average length of the input vector over all the datasets}
				\label{figure:next-activity-input-vector-length}
			\end{figure}
		\end{minipage}
	\end{minipage}
	\begin{minipage}{\linewidth}
		\centering
		\begin{minipage}{0.45\linewidth}
			\begin{figure}[H]
				\includegraphics[trim={2cm 10.2cm 2cm 11.6cm},clip,width=0.9\linewidth]{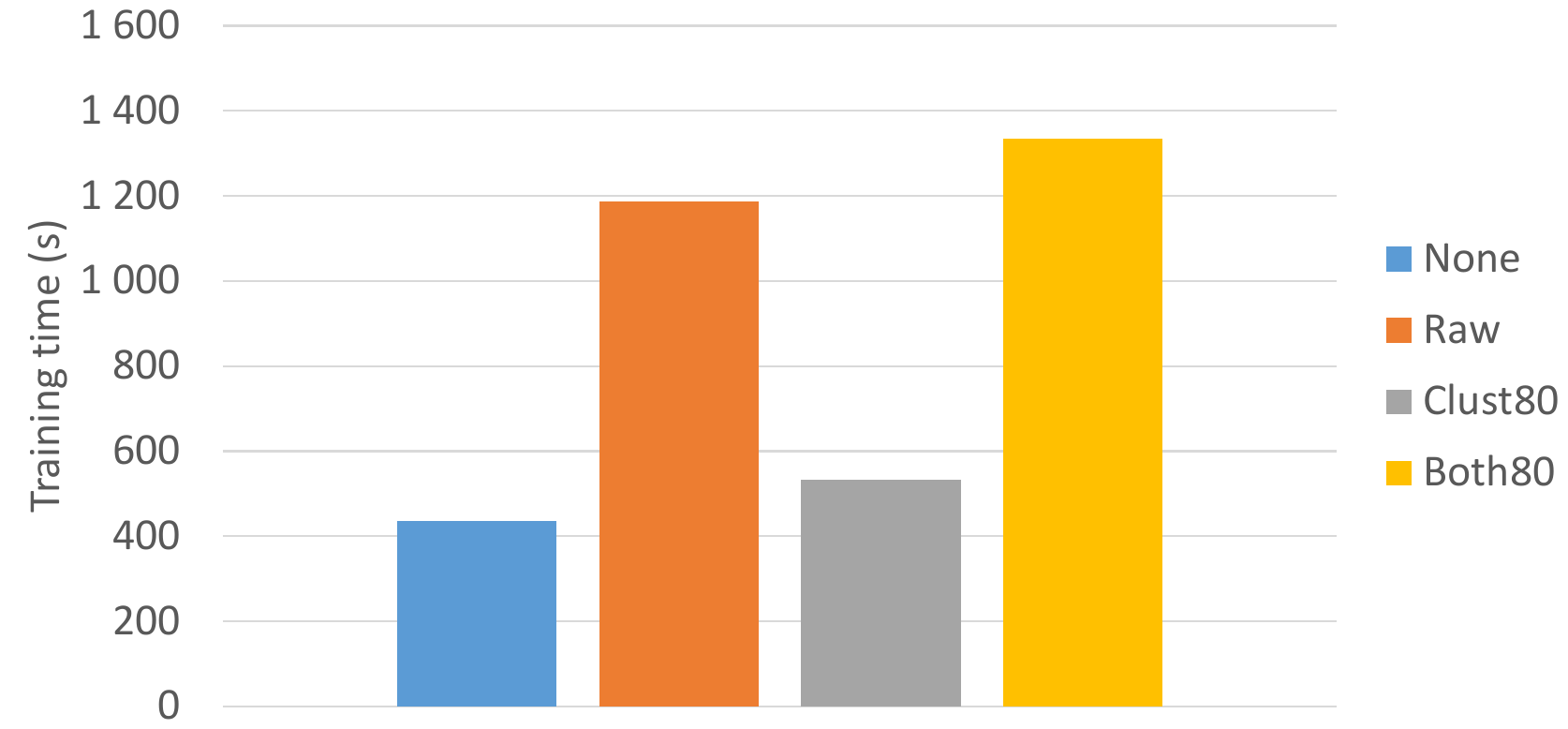}
				\caption[]{Average training time over all the datasets}
				\label{figure:next-activity-training-time}
			\end{figure}
		\end{minipage}
		\hspace{0.05\linewidth}
		\begin{minipage}{0.45\linewidth}
			\begin{figure}[H]
				\includegraphics[trim={2cm 10.2cm 2cm 11.6cm},clip,width=0.9\linewidth]{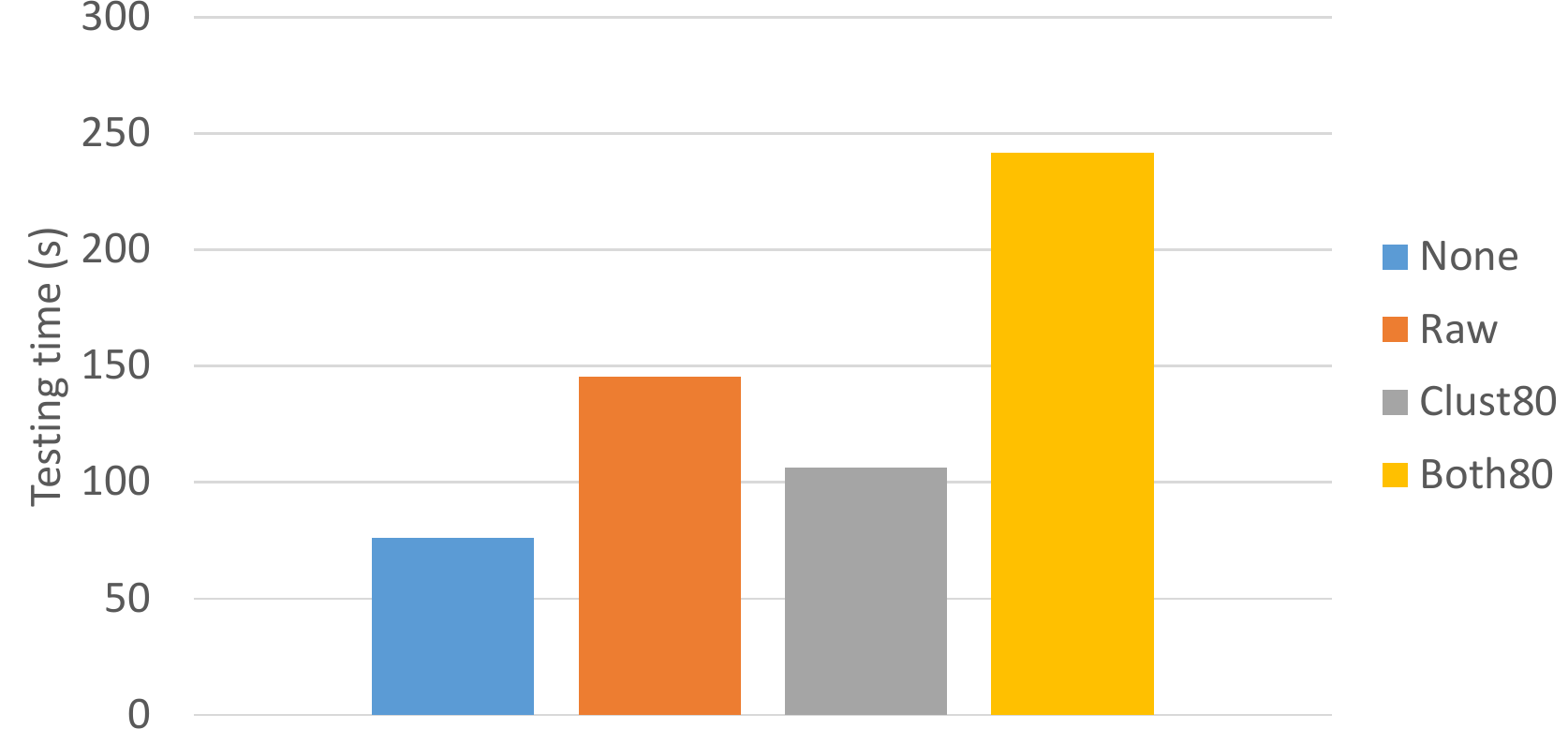}
				\caption[]{Average prediction time over all the datasets}
				\label{figure:next-activity-testing-time}
			\end{figure}
		\end{minipage}
	\end{minipage}
\end{figure*}

\begin{figure*}[!t]
	\begin{minipage}{\linewidth}
		\centering
		\begin{figure}[H]
			\includegraphics[trim={2cm 9.8cm 2cm 11.6cm},clip,width=0.9\linewidth]{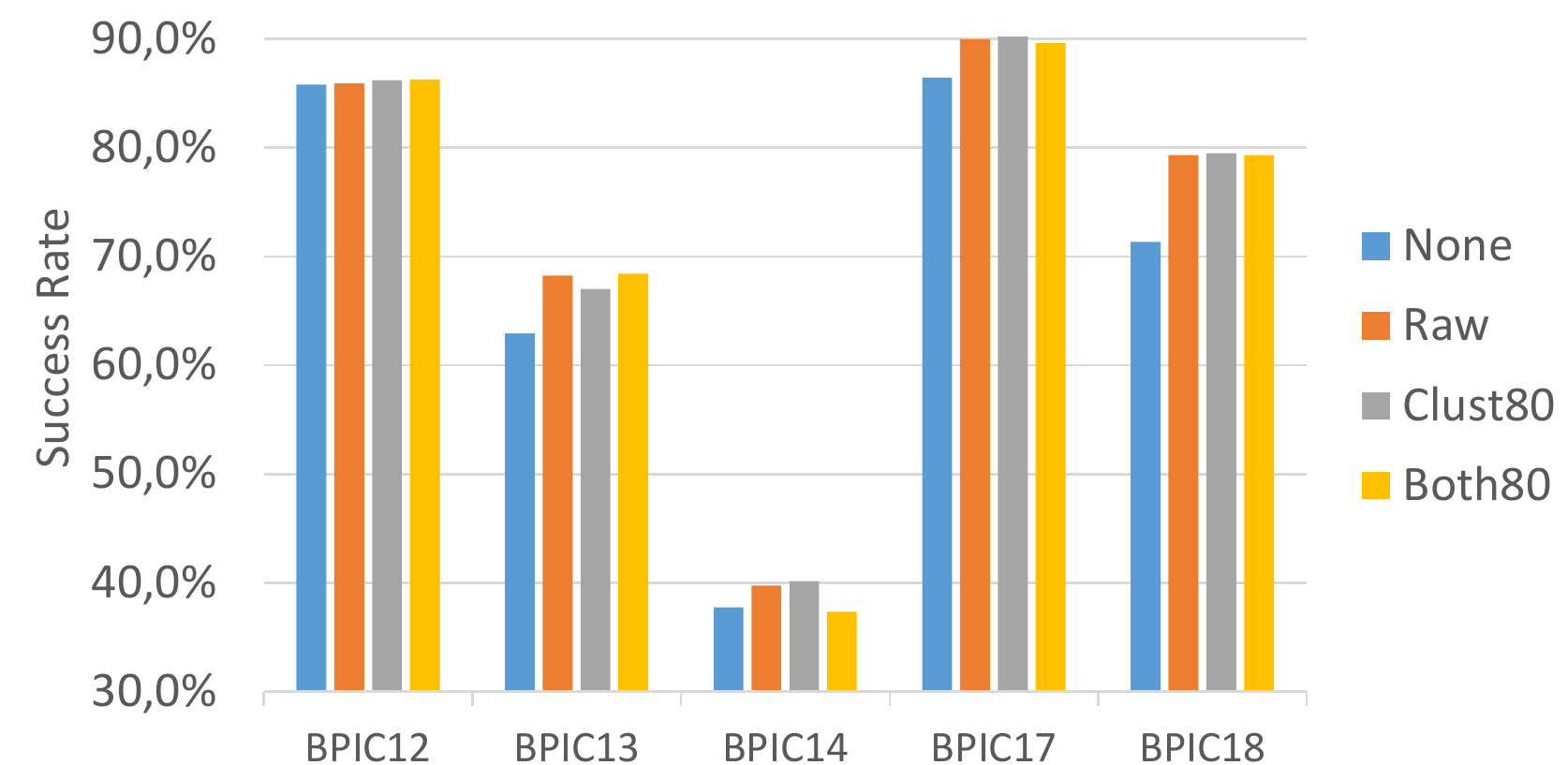}
			\caption[]{Average prediction success rate over all the datasets separately}
			\label{figure:next-activity-training-time-for-each-dataset}
		\end{figure}
	\end{minipage}
\end{figure*}

We also aggregated some of these results over all the datasets using the maximum cluster size of 80 clusters. Figure~\ref{figure:next-activity-success-rate} shows the average success rates of different event attribute encoding techniques over all the tested datasets. Figure~\ref{figure:next-activity-input-vector-length} shows the average input vector lengths. Figure~\ref{figure:next-activity-training-time} and Figure~\ref{figure:next-activity-testing-time} shows the averaged training and prediction times respectively. Finally, Figure~\ref{figure:next-activity-training-time-for-each-dataset} shows the average success rates of different event attribute encoding techniques over all the tested datasets in the case where the maximum cluster count was set to be 80.


Next, we measured the statistical significance of the results. First, we measured whether we could reject the null hypothesis: "the best success rate results could have been achieved by using \textit{Raw} features". By performing a one-tailed $t-test$, while assuming equal variances, for the results, we find out that this hypothesis can not be completely rejected in any of the test datasets. However, when assessing the null hypothesis: "we can achieve better success rates without taking event attributes into account at all than by taking them into account as clustered attribute values", we can reject it in all the datasets. Similarly, we can easily reject null hypothesis: "training a model using \textit{Raw} features can be as fast as using the most accurate tested clustered features" in all the other cases, except in BPIC12, where the used input vectors were essentially identical due to the small number of unique attribute values in that model.


Based on all of the test results and statistical significance analysis, we can see that having event attribute values included improved the prediction accuracy over not having them included at all in all datasets. The effect ranged from 0.5\% in BPIC12 model to 8.5\% in BPIC18. As shown in Figure~\ref{figure:next-activity-success-rate}, very similar success rates were achieved using \textit{ClustN} features as with \textit{Raw}. However, model training and the actual prediction can be performed faster using \textit{ClustN} approaches than either \textit{Raw} or \textit{BothN}. This effect is the most prominently visible in BPIC13 results, where, due to the model having a large amount of unique attribute values, the size of the input vector is almost 68 times bigger and the training time almost 14 times longer using \textit{Raw} feature than \textit{Clust20}. At the same time, the accuracy is still better than not having event attributes at all (about 3.9\% better) and only slightly worse (about 1.4\%) than when using \textit{Raw} feature. This indicates that clustering can be a really powerful technique for minimizing the time required for training especially when there are a lot of unique event attribute values in the used event log. Even when using the maximum cluster count of 20, prediction results will be either not affected or improved with a relatively small impact on the training and prediction time. Input vector size and memory usage are affected by clustered features in a similar fashion with the training and testing time: The more there are unique attribute values in the event log, the greater the difference between the input vector sizes needed in \textit{ClustN} and \textit{Raw} features.

In all the datasets, the best prediction accuracy is always achieved either by using only clustering or by using both clustering and raw attributes at the same time. 

\subsection{Threats to validity}

As threats to the validity of the results in this paper, it is clear that there are a lot of variables involved. As the initial set of parameter values, we used parameters that were found good enough in our earlier work and did some improvement attempts based on the results we got. It is most probable that the set of parameters we used were not optimal ones in each test run. We also did not test all the parameter combinations and the ones we did, we tested often only once, even though there was some randomness involved, e.g., selecting the initial cluster centers in the XMeans algorithm. However, we think that since we tested the results in several different datasets using a 3-fold cross-validation technique, our results can be used at least as a baseline for further studies. All the results generated by the test runs, as well as all the source data and the test framework itself, are available in support materials~\footnote{\url{https://github.com/mhinkka/articles}}. 

Also, we did not test with datasets having a huge number of event attribute values, the maximum amount tested being 2890. However, it can be seen that since the size of the input vectors is completely user-configurable when performing event attribute clustering, the user him/herself can easily set limits to the input vector length which should take the burden off from the RNN and move the burden to the clustering algorithms, which are usually more efficient in handling lots of features and feature values. When evaluating the results of the performed tests and comparing them with other similar works, it should be taken into account that data sampling was used in several phases of the testing process.

\section{Conclusions}
\label{conclusions}

Clustering can be applied to attribute values to improve the accuracy of predictions performed on running cases. In four of the five experimented data sets, having event attribute clusters encoded into the input vectors outperforms having the actual attribute values in the input vector. Also, due to raw attribute values having direct effect to input vector lengths, the training and prediction time will be directly affected by the number of unique event attribute values. Clustering does not have this problem: The number of elements reserved in the input vector for clustered event attribute values can be adjusted freely. The memory usage is directly affected by the length of the input vector. In the tested cases, the number of clusters to use to get the best prediction accuracy seemed to depend very much on the used datasets, when the tested cluster sizes were 20, 40 and 80. In some cases, having more clusters improved the performance, whereas, in others, it did not have any significant impact, or even made the accuracy worse. We also found out that in some cases, having attribute cluster indicators in the input vectors improved the prediction even if the input vectors also included all the actual attribute values. 

As future work, it would be interesting to test this clustering approach also with other machine learning model types such as more traditional random forest and gradient boosting machines. Similarly, it could be interesting to first filter out some of the most rarely occurring attribute values before clustering the values. This could potentially reduce the amount of noise added to the clustered data and make it easier for the clustering algorithm to not be affected by noisy data. Another idea that we leave for future study is whether it would be a good idea to first perform some kind of a feature selection algorithm such as influence analysis~\cite{DBLP:conf/bpm/LehtoHH16}, recursive feature elimination~\cite{granitto2006recursive} or mRMR~\cite{ding2005minimum} to find the attribute values that correlate the most with the prediction results and have those attribute values added into the input vectors as raw one-hot encoded attribute values in addition to having the one-hot encoded cluster labels. More work is also required to understand exactly what properties of the event log affect the optimal number of clusters to use. Finally, more study is required to understand whether a similar clustering approach performed for event attributes in this work could be applicable also for encoding case attributes.

\section{Acknowledgments}
\label{acknowledgments}
We want to thank QPR Software Plc for funding our research. Financial support of Academy of Finland project 313469 is acknowledged.

\label{references}
\bibliography{paper}

\end{document}